\documentclass{bmvc2k}

\usepackage{hyperref}
\usepackage{amssymb}
\usepackage{graphicx}
\usepackage{array}
\usepackage{booktabs} % To thicken table lines
\usepackage{pdfpages}
\usepackage{diagbox}

\graphicspath{ {./images/} }

\newcolumntype{H}{>{\setbox0=\hbox\bgroup}c<{\egroup}@{}}

\makeatletter
\newcommand*{\textlabel}[2]{%
  \edef\@currentlabel{#1}% Set target label
  \phantomsection% Correct hyper reference link
  #1\label{#2}% Print and store label
}
\makeatother

\makeatletter
\newcommand\footnoteref[1]{\protected@xdef\@thefnmark{\ref{#1}}\@footnotemark}
\makeatother

\newcommand\blfootnote[1]{%
  \begingroup
  \renewcommand\thefootnote{}\footnote{#1}%
  \addtocounter{footnote}{-1}%
  \endgroup
}

\title{3D-RETR: End-to-End Single and Multi-View 3D Reconstruction with Transformers}

\addauthor{Zai Shi*}{zaishi@ethz.ch}{1}
\addauthor{Zhao Meng*}{zhmeng@ethz      .ch}{1}
\addauthor{Yiran Xing}{yiran.xing@rwth-aachen.de}{2}
\addauthor{Yunpu Ma}{cognitive.yunpu@gmail.com}{3}
\addauthor{Roger Wattenhofer}{wattenhofer@ethz.ch}{1}

\addinstitution{
 ETH Zurich
}
\addinstitution{
RWTH Aachen
}
\addinstitution{
LMU Munich
}

\runninghead{Shi, Meng, Xing, Ma and Wattenhofer}{3D-RETR}

\begin{document}

\maketitle
\blfootnote{\hspace{-0.55cm} *Equal contribution.}

\begin{abstract}

\end{abstract}

3D reconstruction aims to reconstruct 3D objects from 2D views. Previous works for 3D reconstruction mainly focus on feature matching between views or using CNNs as backbones. Recently, Transformers have been shown effective in multiple applications of computer vision. However, whether or not Transformers can be used for 3D reconstruction is still unclear. In this paper, we fill this gap by proposing 3D-RETR, which is able to perform end-to-end \textbf{3D} \textbf{RE}construction with \textbf{TR}ansformers.  3D-RETR first uses a pretrained Transformer to extract visual features from 2D input images. 3D-RETR then uses another Transformer Decoder to obtain the voxel features. A CNN Decoder then takes as input the voxel features to obtain the reconstructed objects. 3D-RETR is capable of 3D reconstruction from a single view or multiple views. Experimental results on two datasets show that 3D-RETR reaches state-of-the-art performance on 3D reconstruction. Additional ablation study also demonstrates that 3D-DETR benefits from using Transformers.

%-------------------------------------------------------------------------
\section{Introduction}

3D reconstruction focuses on using a single or multiple 2D images of an object to rebuild its 3D representations. 3D reconstruction has played an important role in various downstream applications, including CAD~\cite{cad}, human detection~\cite{human}, architecture~\cite{archi}, etc. The wide applications of 3D reconstruction have motivated researchers to develop numerous methods for 3D reconstruction. Early works for 3D reconstruction mostly use feature matching between different views of an object~\cite{slam, stereoscan, fast_matching}. However, the performance of such methods largely depends on accurate and consistent margins between different views of objects and are thus vulnerable to rapid changes between views~\cite{match_0, match_1, match_2}. Additionally, these methods are not suitable for single-view 3D reconstruction, where only one view of an object is available. 

The advances of deep learning have shed some light on neural network-based approaches for 3D reconstruction~\cite{3d_survey}. On the one hand, some researchers formulate 3D reconstruction as a sequence learning problem and use recurrent neural networks to solve the problem~\cite{3dr2n2, lsm}. On the other hand, other researchers employ the encoder-decoder architecture for 3D reconstruction~\cite{mvcon, pix2vox}. Furthermore, researchers have also used Generative Adversarial Networks (GANs) for 3D reconstruction~\cite{gal}. However, these approaches often rely on sophisticated pipelines of convolutional neural networks (CNNs), and build models with large amounts of parameters, which are computationally expensive. 

Recently, Transformers~\cite{transformers} have gained attention from the computer vision community. Transformer-based models have achieved state-of-the-art performance in many downstream applications of computer vision, including image classification~\cite{vit}, semantic segmentation~\cite{swin}, image super-resolution~\cite{texture}, etc. Despite these achievements, whether or not Transformers can be used in 3D reconstruction is still unclear. 

In this paper, we propose 3D-RETR\footnote{Code: \url{https://github.com/FomalhautB/3D-RETR}}, which is capable of performing end-to-end single and multi-view \textbf{3D} \textbf{RE}construction with \textbf{TR}ansformers. 3D-RETR uses a pretrained Transformer to extract visual features from 2D images. 3D-RETR then obtains the 3D voxel features by using another Transformer Decoder. Finally, a CNN Decoder outputs the 3D representation from the voxel features. Our contributions in this paper are three-folded:

\begin{itemize}
    \item[1.] We propose 3D-RETR for end-to-end single and multi-view 3D reconstruction with Transformers. To the best of our knowledge, we are the first to use Transformers for end-to-end 3D reconstruction. Experimental results show that 3D-RETR reaches state-of-the-art performance under both synthetic and real-world settings. 
    
    \item[2.] We conduct additional ablation studies to understand how each part of 3D-RETR contributes to the final performance. The experimental results show that our choices of the encoder, decoder, and loss are beneficial.

    \item[3.] 3D-RETR is efficient compared to previous models. 3D-RETR reaches higher performance than previous models, despite that it uses far fewer parameters.
    
\end{itemize}

\section{Related Work}

In this Section, we briefly review previous works. Section~\ref{related:3d} gives an overview of previous works on 3D reconstruction. Section~\ref{related:transformers} introduces Transformers.

\subsection{3D reconstruction}\label{related:3d}

3D reconstruction has been widely used in various downstream applications, including architecture~\cite{archi}, CAD~\cite{cad}, human detection~\cite{human}, etc. 
Researchers have mainly focused on two types of methods for 3D reconstruction. Some researchers use depth cameras such as Kinect to collect images with depth information~\cite{depth}, which is subsequently processed for 3D reconstruction. However, such methods require sophisticated hardware and data collection procedures and are thus not practical in many scenarios.

To mitigate this problem, other researchers have resorted to 3D reconstruction from single or multiple views, where only 2D images are available. Early researchers leverage feature matching between different views for 3D reconstruction with 2D images. For example,  \cite{3d_match_0} uses a multi-stage parallel matching algorithm for feature matching, and \cite{fast_matching} proposes a cascade hashing strategy for efficient image matching and 3D reconstruction. Although these methods are useful, their performance degrades when margins between different views are large, making these methods hard to generalize. 

Recent works mainly focus on neural network-based approaches. Some researchers have formulated multi-view 3D construction as a sequence learning problem. For example, \cite{3dr2n2} proposes a 3D recurrent neural network, which takes as input one view at each timestep and outputs the reconstructed object representation. Others employ an encoder-decoder architecture by first encoding the 2D images into fixed-size vectors, from which a decoder decodes the 3D representations~\cite{mvcon, pix2vox, mvsuper}. Furthermore, researchers have also used Generative Adversarial Networks (GANs)~\cite{gal} and 3D-VAEs~\cite{3d_vae_0, 3d_vae_1} for 3D reconstruction. However, these neural network-based methods often rely on sophisticated pipelines of different convolutional neural networks, and are often with models of large amounts of parameters, which are computationally expensive.

\subsection{Transformers}\label{related:transformers}

Researchers first propose Transformers for applications in natural language processing~\cite{transformers}, including machine translation, language modeling, etc. Transformers use a multi-head self-attention mechanism, in which inputs from a specific time step would attend to the entire input sequence. 

Recently, Transformers have also gained attention from the computer vision community. In image classification, Vision Transformer (ViT)~\cite{vit} reach state-of-the-art performance on image classification by feeding images as patches into a Transformer. DeiT~\cite{deit} achieves better performance than ViT~\cite{vit} with much less pretraining data and a smaller parameter size. Transformers are also useful in other computer vision applications. For example, DETR~\cite{detr}, consisting of a Transformer Encoder and a Transformer Decoder, has reached state-of-the-art performance on object detection. Other applications of Transformers also include super-resolution~\cite{texture}, semantic segmentation~\cite{swin}, video understanding~\cite{video}, etc.

In this paper, we propose 3D-RETR for end-to-end single and multi-view 3D reconstruction. 3D-RETR consists of a Transformer Encoder, a Transformer Decoder, and another CNN Decoder. We show in our experiments (see Section~\ref{sec:exp}) that 3D-RETR reaches state-of-the-art performance, while using much fewer parameters than previous models, including pix2vox++~\cite{pix2vox++}, 3D-R2N2~\cite{3dr2n2}, etc.

\subsection{Differentiable Rendering}

Recently, differentiable rendering methods like SRN~\cite{srn}, DVR~\cite{dvr}, NeRF~\cite{nerf}, and IDR~\cite{idr} have become popular. These methods implicitly represent the scene using deep neural networks and have achieved impressive results. However, these methods need to evaluate their neural network thousands of times to extract geometry, which results in a long inference time.

In contrast, our method, along with other previous 3D reconstruction methods including 3D-R2N2~\cite{3dr2n2}, OGN~\cite{ogn} and pix2vox~\cite{pix2vox}, aims to reconstruct the volume without rendering the 2D images. 3D-RETR learns the 3D shape prior out of input 2D images and generates 3D-voxels during the inference time.

\section{Methodology}

\begin{figure*}[!t]
\begin{center}
\includegraphics[width=0.9\linewidth]{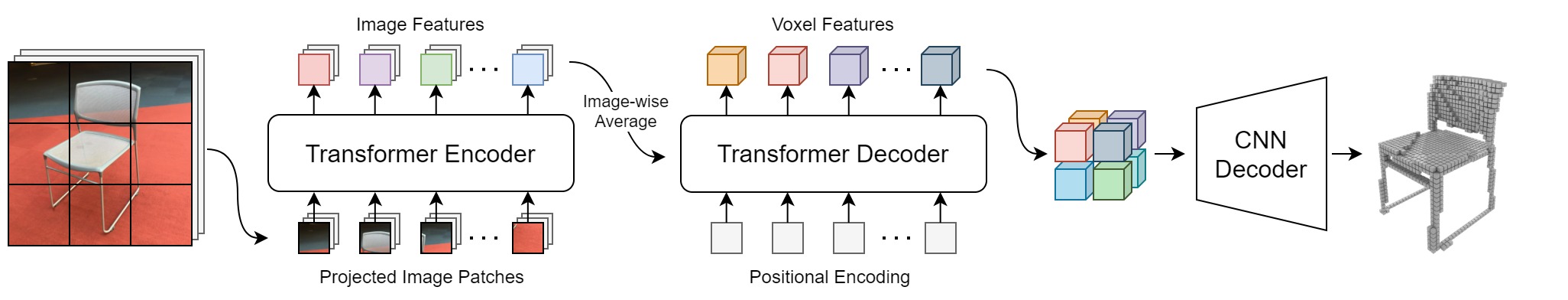}
\end{center}
  \caption{An illustration of our 3D-RETR. A Transformer Encoder firsts extract image features from 2D images. 3D-RETR then obtain the voxel features by using another Transformer Decoder. A CNN Decoder finally outputs the 3D object representation.}
\label{fig:structure}
\end{figure*}

From a high level, 3D-RETR consists of three main components (see Figure \ref{fig:structure}): a Transformer Encoder, a Transformer Decoder, and a CNN Decoder. The Transformer Encoder takes as input the images, which are subsequently encoded into fixed-size image feature vectors. Then, the Transformer Decoder obtains voxel features by cross-attending to the image features. 
Finally, the CNN Decoder decodes 3D object representations from the voxel features. Figure~\ref{fig:structure} illustrates the architecture of 3D-RETR.

In this paper, we have two variants of 3D-RETR: (1) The base model, 3D-RETR-B, has 163M parameters; (2) The smaller model, 3D-RETR-S, has 11M parameters. We describe the details of these two models in Section~\ref{sec:exp}.

We denote the input images of an object from $V$ different views as $x_1, \ldots, x_V \in \mathbb{R}^{C \times H \times W}$, where $C=3$ is the RGB channel, and $H$ and $W$ are the height and width of the images, respectively. We denote the reconstructed voxel by $Y = \{y_1, y_2, \cdots, y_{N^3}| y_i \in \{0, 1\}, 1 \leq i \leq N^3 \}$, 
% $y \in \{0, 1\}^{N^3}$, 
where $i$ is the index to the voxel grids, $0$ indicates an empty voxel grid, $1$ indicates an occupied grid, and $N$ is the resolution of the voxel representation.

\subsection{Transformer Encoder}

A Vision Transformer takes as input image $x_i$ by splitting the image into $B^2$ patches. At each time step, the corresponding patch is embedded by first linearly transformed into a fixed-size vector, which is then added with positional embeddings. The Transformer takes the embedded patch feature as input and outputs $B^2$ encoded dense image feature vectors. For single-view reconstruction, we keep all the $B^2$ dense image vectors. For multi-view reconstruction, at each time step, we take the average across different views, and keep the averaged $B^2$ dense vectors.

In our implementation, we use the Data-efficient image Transformer (DeiT)~\cite{deit}. Our base model, 3D-RETR-B, uses the DeiT Base (DeiT-B) as the Transformer Encoder. DeiT-B consists of 12 layers, each of which has 12 heads and 768-dimensional hidden embeddings. The smaller model, 3D-RETR-S, uses the DeiT Tiny (DeiT-Ti) as the Transformer Encoder. DeiT-Ti has 12 layers, each of which has 3 heads and 192-dimensional hidden embeddings. Both 3D-RETR-B and 3D-RETR-S have $B=16$. We feed all $B^2$ dense image vectors in the next stage into the Transformer Decoder, which we introduce in Section~\ref{sec:transformer_decoder}.

\subsection{Transformer Decoder}\label{sec:transformer_decoder}

The Transformer Decoder takes $M^3$ learned positional embeddings as its input and cross-attends to the output of the Transformer Encoder. Our Transformer Decoder is similar to that of DETR~\cite{detr}, where the Transformer decodes all input vectors in parallel, instead of autoregressively as in the original Transformer~\cite{transformers}.

The 3D-RETR-B model has a Transformer Decoder of 8 layers, each of which has 12 heads and 768-dimensional hidden embeddings. For the 3D-RETR-S, we use a Transformer Decoder of $6$ layers, each of which 3 heads and 192-dimensional hidden embeddings. To enable the Transformer Decoder to understand the spatial relations between voxel features, we create $M^3$ positional embeddings for the Transformer Decoder. The positional embeddings are learnable and are updated during training. We use $M=4$ for both 3D-RETR-B and 3D-RETR-S.

\subsection{CNN Decoder}

\begin{figure*}[!t]
\begin{center}
\includegraphics[width=0.85\linewidth]{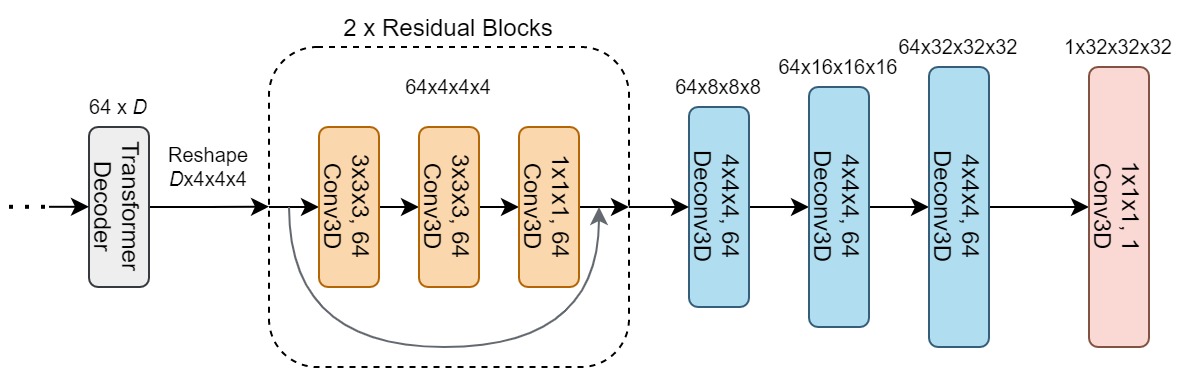}
\end{center}
  \caption{Details of the CNN Decoder in 3D-RETR. The CNN Decoder consists of two residual blocks and three transposed 3D convolutional layers. $D$ is the hidden size of the Transformer.}
\label{fig:cnn_structure}
\end{figure*}

The CNN Decoder takes as input the voxel features from the Transformer Decoder and outputs the voxel representation. As the Transformer Encoder and Transformer Decoder already give rich information, we use a relatively simple architecture for the CNN Decoder. The CNN Decoder first stacks the voxel feature vectors into a cube of size $M^3$, and then upsample the cube iteratively until the desired resolution is obtained.

Figure~\ref{fig:cnn_structure} illustrates the architecture of our CNN Decoder.
Specifically, the CNN Decoder has two residual blocks~\cite{resnet}, each consisting of four transposed 3D convolutional layers. For the residual blocks, the first two convolutional layers have a kernel size of 3, and the last one uses a kernel size of 1. In addition, all three layers have 64 channels. For the transposed 3D convolutional layers, all three layers have a kernel size of 4, a stride of 2, a channel size of 64, and a padding size of 1. 
We add an additional $1\times 1$ convolutional layer at the end of the CNN Decoder to compress the 64 channels into one channel. The model finally outputs cubes of size $32^3$.

\subsection{Loss Function}

While previous works on 3D reconstruction mainly use the cross-entropy loss, researchers have also shown that other losses such as Dice and Focal loss are better for optimizing IoUs~\cite{dice, focal, lovasz}. Although these losses are originally proposed for 2D image tasks, they can be easily adapted to 3D tasks.

This paper uses Dice loss as the loss function, which is suitable for 3D reconstruction as the voxel occupancy is highly unbalanced. Formally, we have the Dice loss for the 3D voxels as follows:
\begin{equation*}
    \mathcal{L}_{Dice} = 1 - \frac{\sum_{n=1}^{N^3} p_n y_n}{\sum_{n=1}^{N^3} p_n + y_n} - \frac{\sum_{n=1}^{N^3} (1-p_n)(1-y_n)}{\sum_{n=1}^{N^3} 2 - p_n - y_n}
\end{equation*}

\noindent where $p_n$ is the $n$-th predicted probability of the voxel occupation and $y_n$ is the $n$-th ground-truth voxel.

\subsection{Optimization}

To train the 3D-RETR. We use the AdamW~\cite{adamw} optimizer with a learning rate of $1e-4$, $\beta_1=0.9$, $\beta_2=0.999$, and a weight decay of $1e-2$. The batch size is set to 16 for all the experiments. We use two RTX Titan GPUs in our experiments. Training takes 1 to 3 days, depending on the exact setting. We use mixed-precision to speed up training.

\section{Experiments}\label{sec:exp}

We show in this Section our experimental results. We evaluate 3D-RETR on ShapeNet~\cite{shapenet} and Pix3d~\cite{pix3d}. Following previous works~\cite{3dr2n2, pix2vox}, we use Intersection of Union (IoU) as our evaluation metric.

\subsection{ShapeNet}

ShapeNet is a large-scale 3D object dataset consisting of $55$  object categories with $51,300$ 3D models. Following the setting in Pix2Vox~\cite{pix2vox}, we use the same subset of $13$ categories and about $44,000$ models. The 3D models are pre-processed using Binvox\footnote{\url{https://www.patrickmin.com/binvox/}} with a resolution of $32^3$.\footnote{We asked the authors of previous works for higher resolution datasets.  Unfortunately, the authors do not have access to the datasets anymore. Therefore, we cannot compare and evaluate the performance of other resolutions.} The images are then rendered in the resolution of $137 \times 137$ from $24$ random views. 

For single-view 3D reconstruction on ShapeNet, we compare our results with  previous state-of-the-art models, including 3D-R2N2~\cite{3dr2n2}, OGN~\cite{ogn}, Matryoshka Networks~\cite{matryoshka}, AtlasNet~\cite{atlasnet}, Pixel2Mesh~\cite{pixel2mesh}, OccNet~\cite{occnet}, IM-Net~\cite{imnet}, AttSets~\cite{attsets}, and Pix2Vox++~\cite{pix2vox++}. 
Table~\ref{table:single_view} shows the results. We can observe that both 3D-RETR-S and 3D-RETR-B outperform all previous models in terms of overall IoU. Additionally, 3D-RETR-S outperforms all other baselines in 6 of the 13 categories, while 3D-RETR-B is the best among other baselines in 9 of the 13 categories.

\begin{table}[!t]
\begin{center}
\resizebox{0.99\linewidth}{!}{
\begin{tabular}{ccccccccccccc}
\toprule
Category    & 3D-R2N2   & OGN           & Matroyoshka           & AtlasNet              & Pixel2Mesh        & OccNet    & IM-Net        
            & AttSets   & Pix2Vox++     & \textbf{3D-RETR-S}   & \textbf{3D-RETR-B} \\
\midrule\midrule % --------------------------------------------------------------------------------------------------------------------
aeroplane   & 0.513     & 0.587         & 0.647                 & 0.493                 & 0.508         & 0.532         & 0.702         
            & 0.594     & 0.674         & \textit{0.696}                 & \textbf{0.704} \\
% ---------------------------------------------------------------------------------------------------------------------------------
bench       & 0.421     & 0.481         & 0.577                 & 0.431                 & 0.379         & 0.597         & 0.564         
            & 0.552     & 0.608         & \textit{0.643}                 & \textbf{0.650} \\
% ---------------------------------------------------------------------------------------------------------------------------------            
cabinet     & 0.716     & 0.729         & 0.776                 & 0.257                 & 0.732         & 0.674         & 0.680         
            & 0.783     & 0.799         & \textbf{0.804}        & \textit{0.802} \\
% ---------------------------------------------------------------------------------------------------------------------------------
car         & 0.798     & 0.816         & 0.850                 & 0.282                 & 0.670         & 0.671         & 0.756         
            & 0.844     & 0.858         & \textit{0.858}                 & \textbf{0.861} \\
% ---------------------------------------------------------------------------------------------------------------------------------
chair       & 0.466     & 0.483         & 0.547                 & 0.328                 & 0.484         & 0.583         & \textbf{0.644}
            & 0.559     & 0.581         & 0.579                 & \textit{0.592} \\
% ---------------------------------------------------------------------------------------------------------------------------------            
display     & 0.468     & 0.502         & 0.532                 & 0.457                 & 0.582         & \textbf{0.651}& \textit{0.585}         
            & 0.565     & 0.548         & 0.576                 & 0.574 \\ 
% ---------------------------------------------------------------------------------------------------------------------------------            
lamp        & 0.381     & 0.398         & 0.408                 & 0.261                 & 0.399         & \textit{0.474}         & 0.433         
            & 0.445     & 0.457         & 0.463                 & \textbf{0.483} \\
% ---------------------------------------------------------------------------------------------------------------------------------            
rifle       & 0.544     & 0.593         & 0.616                 & 0.573                 & 0.468         & 0.656         & \textbf{0.723}
            & 0.601     & \textit{0.721}         & 0.665                 & 0.668 \\
% ---------------------------------------------------------------------------------------------------------------------------------            
sofa        & 0.628     & 0.646         & 0.681                 & 0.354                 & 0.622         & 0.669         & 0.694         
            & 0.703     & 0.725         & \textit{0.729}                 & \textbf{0.735} \\
% ---------------------------------------------------------------------------------------------------------------------------------            
speaker     & 0.662     & 0.637         & 0.701                 & 0.296                 & 0.672         & 0.655         & 0.683         
            & \textit{0.721}     & 0.617         & 0.719                 & \textbf{0.724} \\
% ---------------------------------------------------------------------------------------------------------------------------------            
table       & 0.513     & 0.536         & 0.573                 & 0.301                 & 0.536         & \textbf{0.659}& 0.621         
            & 0.590     & 0.620         & 0.615                 & \textit{0.633} \\
% ---------------------------------------------------------------------------------------------------------------------------------            
telephone   & 0.661     & 0.702         & 0.756                 & 0.543                 & 0.762         & 0.794         & 0.762         
            & 0.743     & \textbf{0.809}& \textit{0.796}                 & 0.781 \\
% ---------------------------------------------------------------------------------------------------------------------------------            
watercraft  & 0.513     & \textit{0.632}         & 0.591                 & 0.355                 & 0.471         & 0.579         & 0.607         
            & 0.601     & 0.603         & 0.621                 & \textbf{0.636} \\
\midrule % --------------------------------------------------------------------------------------------------------------------------
overall     & 0.560     & 0.596         & 0.635                 & 0.352                 & 0.552         & 0.626         & 0.659         
            & 0.642*    & 0.670*        & \textit{0.674}                 & \textbf{0.680} \\
\bottomrule
\end{tabular}
}
\end{center}
\caption{Results of single-view 3D reconstruction on the ShapeNet dataset. \textbf{Bold} is the best performance, while \textit{Italic} is the second best. For overall IoU, we report the mean IoU across all 13 categories. However, for entries with $*$, the overall IoU is NOT the averaged IoU across categories. We nevertheless use the original number from Pix2Vox++~\cite{pix2vox++} and AttSets~\cite{attsets}. As a reference, the average IoU across categories for AttSets and Pix2Vox++ are $0.638$ and $0.663$, respectively.}
\label{table:single_view}
\end{table}

\begin{figure*}[ht]
\begin{center}
\includegraphics[width=0.6\textwidth]{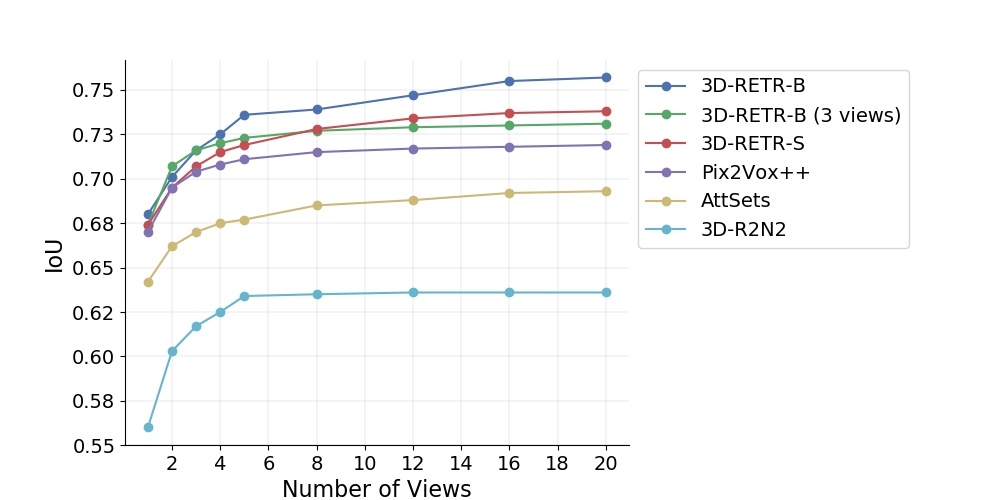}
\end{center}
  \caption{Model performance with different views. 3D-RETR-B continues to benefit from more views, while baselines including AttSets and Pix2Vox++ become saturated.}
\label{fig:multi_view2}
\end{figure*}

For the multi-view setting, we take the number of input 2D images $V \in \{1, 2, 3, 4, 5, 8, 12,\\ 16, 20\}$ and compare the performance of 3D-RETR with previous state-of-the-art models, including 3D-R2N2~\cite{3dr2n2}, AttSets~\cite{attsets}, and Pix2Vox++~\cite{pix2vox++}. As we can see in Table~\ref{table:multi_view}, 3D-RETR outperforms all previous works on all different views. Furthermore,   Figure~\ref{fig:multi_view2} illustrates the relation between the number of views and model performance. One can observe that the performance of 3D-RETR increases rapidly compared to other methods as more views become available. Additionally, while models like 3D-R2N2 and AttSets gradually become saturated, our best model, 3D-RETR-B, continues to benefit from more views, indicating that 3D-RETR has a higher capacity.

As 3D-RETR simply takes the average over different views in the Transformer Encoder, we can train and evaluate 3D-RETR with different numbers of views. To understand how 3D-RETR performs when the number of views is different during training and evaluation, we conduct additional experiments to train 3D-RETR-B with 3 views and evaluate its performance under different numbers of views. We show the results in Table~\ref{table:multi_view} (See the row of 3D-RETR-B (3 views)). Surprisingly, 3D-RETR-B still outperforms previous state-of-the-art models, even if the number of views during training and evaluation is different. In particular, the model seeing different numbers of views during training and evaluation demonstrates that 3D-RETR is flexible.

\begin{table}[!t]
\begin{center}
\resizebox{0.85\linewidth}{!}{
\begin{tabular}{cccccccccc}
\toprule
Model                           & 1 view            & 2 views           & 3 views           & 4 views           & 5 views 
                                & 8 views           & 12 views          & 16 views          & 20 views  \\
\midrule\midrule
3D-R2N2                         & 0.560             & 0.603             & 0.617             & 0.625             & 0.634
                                & 0.635             & 0.636             & 0.636             & 0.636     \\
AttSets                         & 0.642             & 0.662             & 0.670             & 0.675             & 0.677
                                & 0.685             & 0.688             & 0.692             & 0.693     \\
Pix2Vox++*                      & 0.670             & 0.695             & 0.704             & 0.708             & 0.711
                                & 0.715             & 0.717             & 0.718             & 0.719     \\ \midrule
\textbf{3D-RETR-S}             & 0.674             & 0.695             & 0.707             & 0.715             & 0.719
                                & 0.728             & 0.734             & 0.737             & 0.738     \\
\textbf{3D-RETR-B} (3 views)   & 0.674             & \textbf{0.707}    & \textbf{0.716}    & 0.720             & 0.723
                                & 0.727             & 0.729             & 0.730             & 0.731     \\
\textbf{3D-RETR-B}             & \textbf{0.680}    & 0.701             & \textbf{0.716}    & \textbf{0.725}    & \textbf{0.736}      
                                & \textbf{0.739}    & \textbf{0.747}    & \textbf{0.755}    & \textbf{0.757} \\
 
\bottomrule
\end{tabular}
}
\end{center}
\caption{Results of multi-view 3D reconstruction on the ShapeNet dataset. Our smallest model (3D-RETR-S) already reaches state-of-the-art performance. *: As mentioned in Table~\ref{table:single_view}, while all other models report mean IoU across categories, Pix2Vox++~\cite{pix2vox++} reports their overall IoU by taking the average across all the examples. For Pix2Vox++, we cannot compute mean IoU across different categories as Pix2Vox++ does not report per-category IoU for multi-view reconstruction.}
\label{table:multi_view}
\end{table}

\newcommand\w{3cm}
\newcommand{\samplerow}[1]{
\includegraphics[width=\w]{images/sample#1/input.jpg} & 
\includegraphics[width=\w]{images/sample#1/3D-R2N2.jpg} & 
\includegraphics[width=\w]{images/sample#1/Atlasnet.jpg} & 
\includegraphics[width=\w]{images/sample#1/OccNet.jpg} &
\includegraphics[width=\w]{images/sample#1/IM-NET.jpg} &
\includegraphics[width=\w]{images/sample#1/AttSets.jpg} &
\includegraphics[width=\w]{images/sample#1/Pix2Vox++F.jpg} & 
\includegraphics[width=\w]{images/sample#1/Voxel-Transformer-S.jpg} & 
\includegraphics[width=\w]{images/sample#1/Voxel-Transformer.jpg} &
\includegraphics[width=\w]{images/sample#1/gt.jpg} \\
}

\begin{table}[!t]
\Large
\begin{center}
\resizebox{0.95\linewidth}{!}{
\begin{tabular}{ccccccccccc}
\toprule
Input & 3D-R2N2 & AtlasNet & OccNet & IM-NET & AttSets& Pix2Vox++ & \textbf{3D-RETR-S} & \textbf{3D-RETR-B} & GT \\
\midrule\midrule 
\samplerow{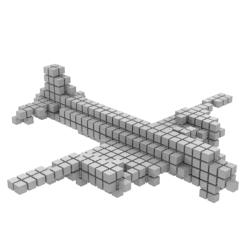}
\samplerow{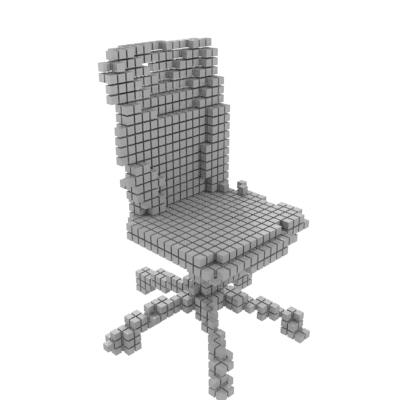}
\samplerow{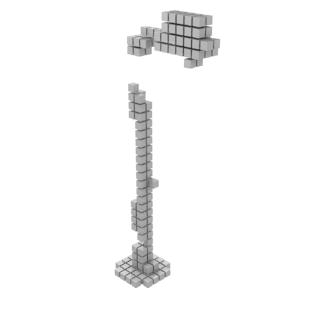}
\samplerow{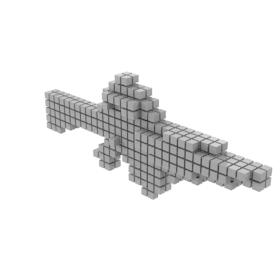}
\samplerow{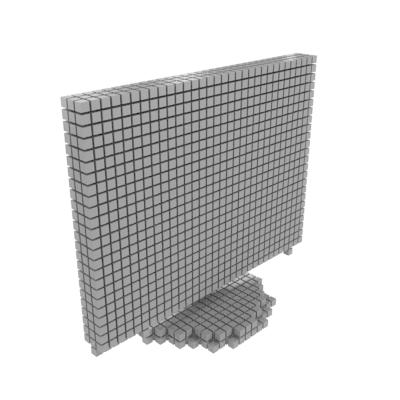}
\bottomrule
\end{tabular}
}
\end{center}
\caption{Examples of single-view 3D reconstruction from the ShapeNet dataset.}
\label{table:shapenet_sample}
\end{table}

\subsection{Pix3D}

Different from ShapeNet, in which all examples are synthetic, 
Pix3D~\cite{pix3d} is a dataset of aligned 3D models and real-world 2D images. Evaluating models on Pix3D gives a better understanding of the model performance under practical settings. Following the same setting in Pix3D~\cite{pix3d}, we use the subset consisting of 2,894 untruncated and unoccluded chair images as the test set. Moreover, we follow~\cite{renderforcnn} to synthesize 60 random images for each image in the ShapeNet-Chair category and use these synthesized images as our training set.

We compare 3D-RETR with previous state-of-art models, including DRC~\cite{drc}, 3D-R2N2~\cite{3dr2n2}, Pix3D~\cite{pix3d}, and Pix2Vox++~\cite{pix2vox++}. Table~\ref{table:pix3d} shows the results. 3D-RETR-B outperforms all previous models, and 3D-RETR-S reaches comparable performance despite that 3D-RETR-S is much smaller than 3D-RETR-B in terms of parameter size.

\begin{table}[!t]
\begin{center}
\resizebox{0.55\linewidth}{!}{
\begin{tabular}{ccccccc}
\toprule
3D-R2N2 & DRC   & Pix3D & Pix2Vox++    & \textbf{3D-RETR-S}   & \textbf{3D-RETR-B} \\
\midrule\midrule
0.136   & 0.265 & 0.282 & 0.288      & 0.283                 & \textbf{0.290} \\
 
\bottomrule
\end{tabular}
}
\end{center}
\caption{Results of single-view reconstruction on the Pix3D-Chair dataset.}
\label{table:pix3d}
\end{table}

\begin{table}[!t]
\Large
\begin{center}
\resizebox{0.46\linewidth}{!}{
\begin{tabular}{cccc}
\toprule
Input & 3D-RETR-S & 3D-RETR-B & GT \\
\midrule\midrule 
\includegraphics[width=3cm,height=3cm,keepaspectratio]{images/pix3d_smaple1/input.jpg} & 
\includegraphics[width=3cm]{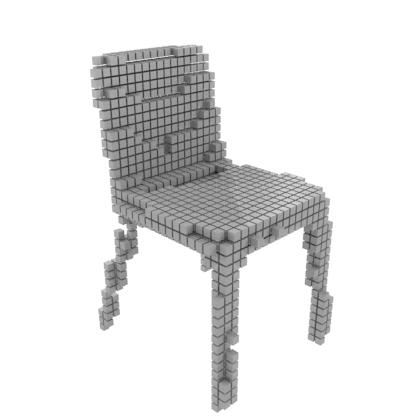} &
\includegraphics[width=3cm]{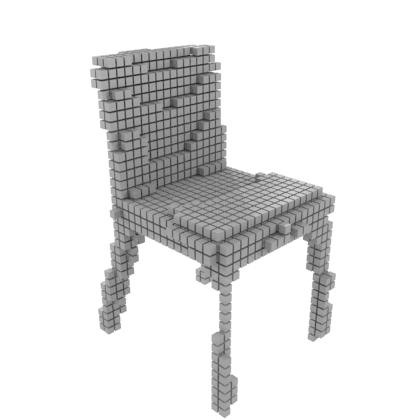} &
\includegraphics[width=3cm]{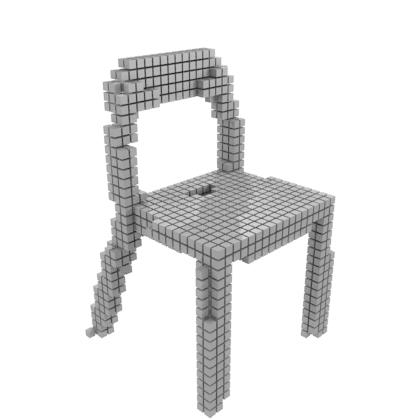}\\

\includegraphics[width=3cm,height=3cm,keepaspectratio]{images/pix3d_smaple2/input.jpg} & 
\includegraphics[width=3cm]{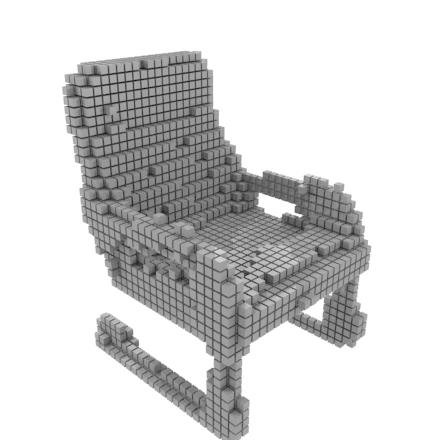} &
\includegraphics[width=3cm]{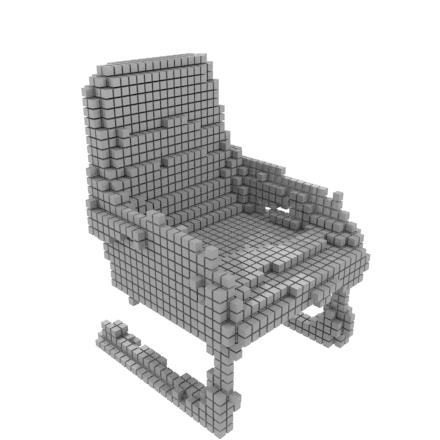} &
\includegraphics[width=3cm]{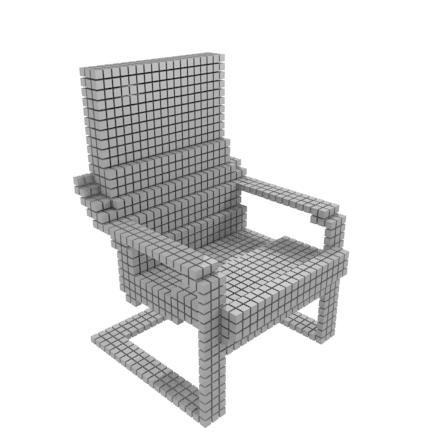}\\

\includegraphics[width=3cm,height=3cm,keepaspectratio]{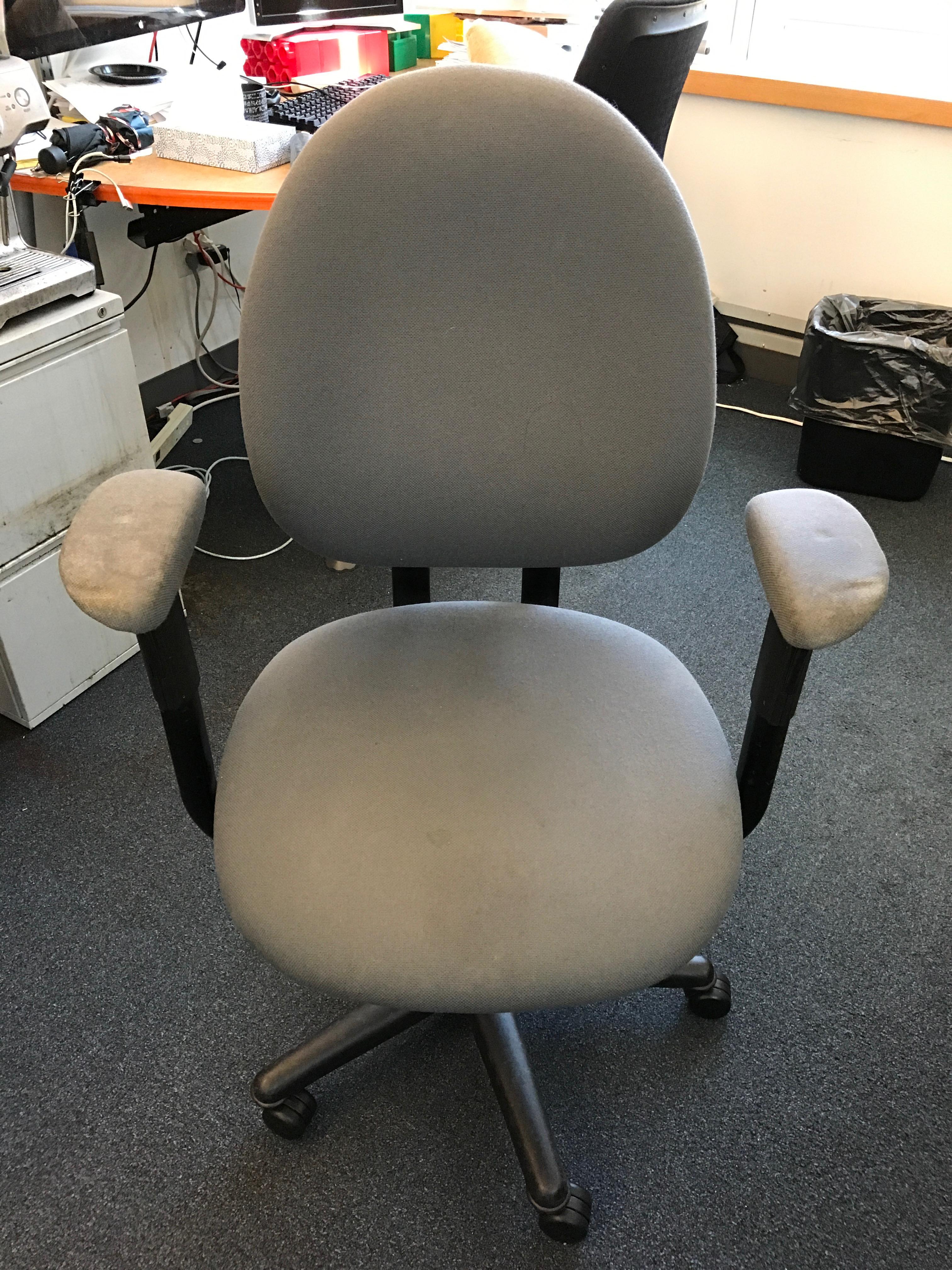} & 
\includegraphics[width=3cm]{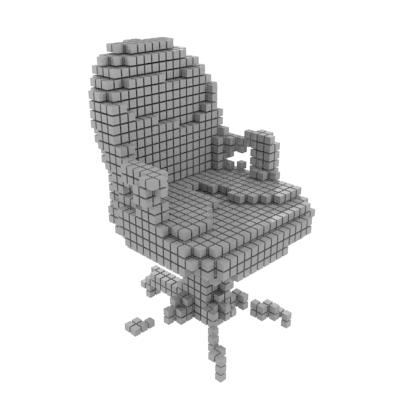} &
\includegraphics[width=3cm]{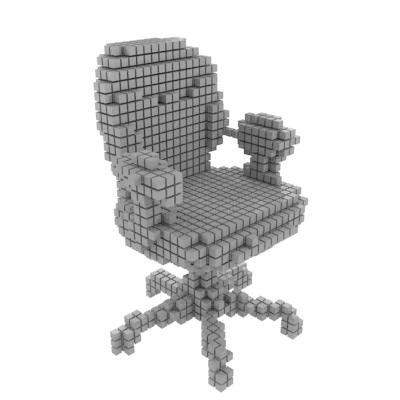} &
\includegraphics[width=3cm]{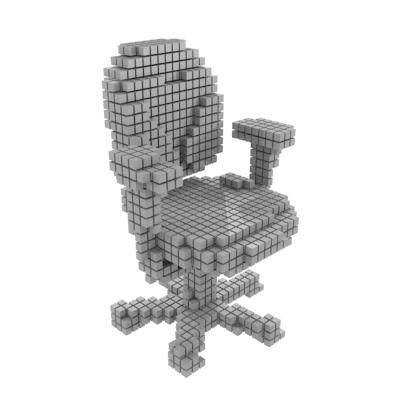}\\
\bottomrule
\end{tabular}
}
\end{center}
\caption{Example outputs of 3D-RETR on single-view 3D reconstruction from the Pix3D dataset. We do not show examples from baseline models as none of the baselines have released their implementation on Pix3D.}
\label{table:pix3d_sample2}
\end{table}

% \section{Discussion}

\subsection{Ablation Study}
\label{ablation}

\begin{table}[!t]
\begin{center}
\resizebox{0.65\linewidth}{!}{
\begin{tabular}{cccccc}
\toprule
Name                        & Encoder           & First Decoder    & Second Decoder    & Loss      & IoU \\
\midrule\midrule
3D-RETR-B                  & Base              & Base          & CNN           & Dice      & \textbf{0.680} \\
3D-RETR-S                  & Small             & Small         & CNN           & Dice      & 0.674 \\
\textlabel{Setup 1}{setup1} & Base              & Tiny          & CNN           & Dice      & 0.667 \\
\textlabel{Setup 2}{setup2} & Base (w/o pre.)   & Base          & CNN           & Dice      & 0.279 \\
\textlabel{Setup 3}{setup3} & ResNet-50         & Base          & CNN           & Dice      & 0.670 \\
\textlabel{Setup 4}{setup4} & Base              & Base          & VQ-VAE        & -         & 0.598 \\
\textlabel{Setup 5}{setup5} & Base              & Base          & CNN           & CE        & 0.668 \\
\textlabel{Setup 6}{setup6} & Base              & Base          & MLP           & Dice      & 0.658 \\

\bottomrule
\end{tabular}
}
\end{center}
\caption{Ablation Study. We ablate 3D-RETR by using different encoders, decoders, and loss functions.}
\label{table:ablation}
\end{table}

We ablate 3D-RETR by using different Transformer Encoders, Transformer Decoders, CNN Decoders, and loss functions. Table~\ref{table:ablation} shows the results of our ablation study. Specifically, we discuss the following model variants: 

\begin{itemize}
    \item \ref{setup1}: One might think that the Transformer Decoder is redundant, as the Transformer Encoder and CNN Decoder are already available. We show that the Transformer Decoder is necessary by replacing it with a tiny Transformer Decoder. The tiny Transformer Decoder has only $1$ layer and $1$ head, which serves only as a simple mapping between the outputs of the Transformer Encoder and the input of the CNN Decoder. We can see that the performance decreases from 0.680 to 0.667 after using the tiny Transformer Decoder.
    
    \item \ref{setup2}: Pretraining for the Transformer Encoder is crucial since Transformers large amounts of data to gain prior knowledge for images. In this setup, we observe that the performance of 3D-RETR decreases significantly without pretraining.
    
    \item \ref{setup3}: We show the advantage of the Transformer Encoder over a CNN Encoder by replacing the Transformer Encoder with a pretrained ResNet-50~\cite{resnet}. After replacing, the model performance decreases from 0.680 to 0.670.
    
    \item \ref{setup4}: Previous studies, including VQ-VAE~\cite{vqvae}, VQ-GAN~\cite{vqgan}, and DALL$\cdot$E~\cite{dalle} have employed a two-stage approach for generating images. We adopt a similar approach to 3D-RETR by first training a 3D VQ-VAE~\cite{vqvae} and replacing the CNN Decoder with the VQ-VAE Decoder. In this setting, the Transformer Decoder decodes autoregressively. The training process for this variant is also different from the standard 3D-RETR. We first generate the discretized features using ground-truth voxels and the VQ-VAE Encoder. These discretized features are then used as the ground truth for the Transformer Decoder. During the evaluation, the Transformer Decoder generates the discretized features one by one and then feeds them into the VQ-VAE Decoder. We show in Table~\ref{table:ablation} that the performance of this two-stage approach is not as good as our single-stage setup.
    
    \item \ref{setup5}: To understand how loss functions affect model performance, we train a 3D-RETR-B with the standard cross-entropy loss. From Table~\ref{table:ablation}, we can see that replacing Dice loss with cross-entropy loss results in performance degradation, indicating that Dice loss is optimal for 3D-RETR.

    \item \ref{setup6}: We replace the CNN Decoder with a simple one-layer MLP, so the model becomes a pure Transformer model. The performance is not as good as the original model with CNN Decoder, but still achieves comparable results.

\end{itemize}

We give further comparisons of parameter size and model performance in Table~\ref{tab:size}.  Despite that our 3D-RETR-S is smaller than previous state-of-the-art models, it still reaches better performance. Furthermore, 3D-RETR-B outperforms 3D-RETR-S, showing that increasing the parameter size is helpful for 3D-RETR.  

\begin{table}[!t]
\begin{center}
\resizebox{0.75\linewidth}{!}{
\begin{tabular}{ccccccc}
\toprule
Model           & 3D-R2N2   & OGN   & Matryoshka    & Pix2Vox++ & \textbf{3D-RETR-S}    & \textbf{3D-RETR-B} \\
\midrule\midrule
\#Parameter     & 36M   & 12M   & 46M           & 98M       & 11M                   & 163M    \\
IoU             & 0.560 & 0.596 & 0.635         & 0.670*    & 0.674                 & 0.680   \\
 
\bottomrule
\end{tabular}
}
\end{center}
\caption{Parameter size and performance comparison between 3D-RETR and other baseline models. 3D-RETR reaches better performance with fewer parameters. *See Table~\ref{table:single_view} and Table~\ref{table:multi_view}.}
\label{tab:size}
\end{table}

\section{Conclusion}

Despite that Transformers have been widely used in various applications in computer vision~\cite{vit, detr, texture}, whether or not Transformers can be used for single and multi-view 3D reconstruction remains unclear. In this paper, we fill in this gap by proposing \textbf{3D-RETR}, which is capable of performing end-to-end single and multi-view \textbf{3D} \textbf{RE}construction with \textbf{TR}ansformers. 3D-RETR consists of a Transformer Encoder, a Transformer Decoder, and a CNN Decoder. Experimental results show that 3D-RETR reaches state-of-the-art performance on 3D reconstruction under both synthetic and real-world settings. 3D-RETR is more efficient than previous models~\cite{pix2vox, ogn, matryoshka}, as 3D-RETR reaches better performance with much fewer parameters. In the future, we plan to improve 3D-RETR by using other variants of Transformers, including Performer~\cite{performer}, Reformer~\cite{reformer}, etc.

\bibliography{citations}

\clearpage
\appendix

\section{3D-RETR with VQ-VAE}

\begin{figure*}[!t]
\begin{center}
\includegraphics[width=0.75\linewidth]{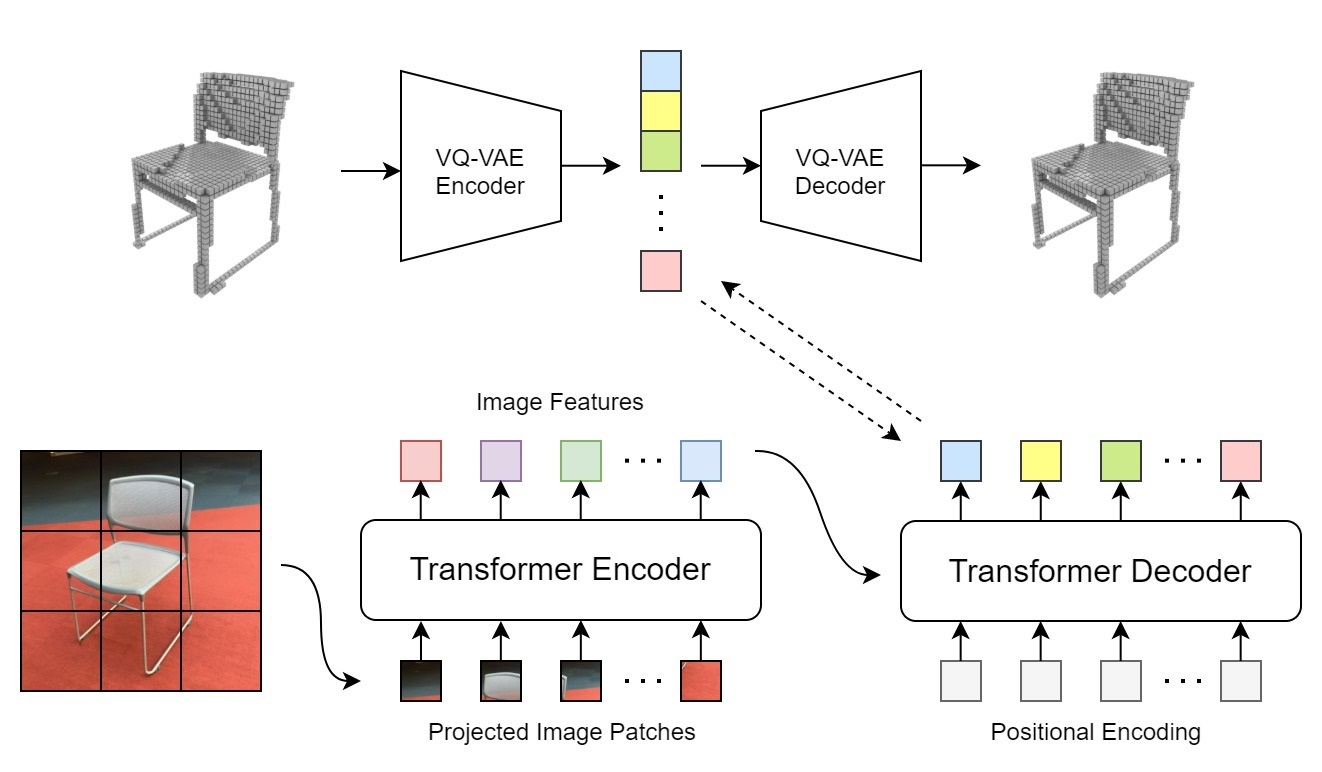}
\end{center}
  \caption{3D-RETR with VQ-VAE. This corresponds to Setup 4 of our ablation study.}
\label{fig:vqvae}
\end{figure*}

We describe in detail the VQ-VAE setting in our ablation study of Section~4.3~(see Figure~\ref{fig:vqvae}). We train 3D-RETR with VQ-VAE in two separate stages.

In the first stage, we pretrain a VQ-VAE with a codebook size of 2048, where each codebook vector has $512$ dimensions. The VQ-VAE Encoder and Decoder have three layers, respectively. For the VQ-VAE Decoder, we use the same residual blocks as in the CNN Decoder. The VQ-VAE Encoder encodes the $32 \times 32 \times 32$ voxel into a discrete sequence of length 64, where each element in the sequence is an integer between 0 and 2047. The VQ-VAE is trained with cross-entropy loss. The reconstruction IoU is about 0.885.

In the second stage, for every input image $x$ and its correspondent ground-truth voxel $Y$, we first generate a discrete sequence $D$ using the pretrained VQ-VAE Encoder. Then, the Transformer Encoder generates the hidden representation for the input image $x$, and the Transformer Decoder uses the output of the Transformer Encoder to generate another discrete sequence $D^\prime$. To generate $D^\prime$, we use a linear layer with softmax at the output of the Transformer Decoder. We use the sequence $D$ as the ground truth and train the Transformer Encoder and Decoder with cross-entropy loss to generate $D^\prime$, which should be as close as possible to $D$.

\section{Additional Examples}

We show more examples of the ShapeNet dataset and the Pix3D dataset from our 3D-RETR-B model. Table~\ref{table:pix3d_sample} shows additional examples of the Pix3D dataset. Table~\ref{table:shapenet_views} shows examples from the ShapeNet dataset with different numbers of views as inputs. We can see a clear quality improvement when more views become available.

\newcommand{\pix}[2]{\raisebox{-.5\height}{\includegraphics[width=3cm,height=3cm,keepaspectratio]{sup_images/pix3d/#1/#2.jpg}}}

\begin{table}[!h]
\Large
\begin{center}
\resizebox{\linewidth}{!}{
\begin{tabular}{cccccc}
\toprule
Input & \pix{4}{input}  & \pix{2}{input}    & \pix{3}{input}    & \pix{1}{input}    & \pix{5}{input}\\
Prediction & \pix{4}{prediction}  & \pix{2}{prediction}    & \pix{3}{prediction}    & \pix{1}{prediction}    & \pix{5}{prediction}\\
GT & \pix{4}{gt}  & \pix{2}{gt}    & \pix{3}{gt}    & \pix{1}{gt}    & \pix{5}{gt}\\
\bottomrule
\end{tabular}
}
\end{center}
\caption{Examples from the Pix3D dataset. All predictions are generated by 3D-RETR-B.}
\label{table:pix3d_sample}
\end{table}

\newcommand{\imgview}[2]{\raisebox{-.5\height}{\includegraphics[width=3cm,height=3cm,keepaspectratio]{sup_images/shapenet_views/#1/#2.jpg}}}

\begin{table}[!t]
\Large
\begin{center}
\resizebox{\linewidth}{!}{
\begin{tabular}{ccccccc}
\toprule
Input           & \imgview{1}{input}    & \imgview{2}{input}& \imgview{3}{input} & \imgview{4}{input} & \imgview{5}{input} & \imgview{6}{input}\\
1 view          & \imgview{1}{1v}       & \imgview{2}{1v}   & \imgview{3}{1v}    & \imgview{4}{1v}    & \imgview{5}{1v}    & \imgview{6}{1v}\\
2 views         & \imgview{1}{2v}       & \imgview{2}{2v}   & \imgview{3}{2v}    & \imgview{4}{2v}    & \imgview{5}{2v}    & \imgview{6}{2v}\\
3 views         & \imgview{1}{3v}       & \imgview{2}{3v}   & \imgview{3}{3v}    & \imgview{4}{3v}    & \imgview{5}{3v}    & \imgview{6}{3v}\\
4 views         & \imgview{1}{4v}       & \imgview{2}{4v}   & \imgview{3}{4v}    & \imgview{4}{4v}    & \imgview{5}{4v}    & \imgview{6}{4v}\\
5 views         & \imgview{1}{5v}       & \imgview{2}{5v}   & \imgview{3}{5v}    & \imgview{4}{5v}    & \imgview{5}{5v}    & \imgview{6}{5v}\\
8 views         & \imgview{1}{8v}       & \imgview{2}{8v}   & \imgview{3}{8v}    & \imgview{4}{8v}    & \imgview{5}{8v}    & \imgview{6}{8v}\\
12 views        & \imgview{1}{12v}      & \imgview{2}{12v}  & \imgview{3}{12v}   & \imgview{4}{12v}   & \imgview{5}{12v}   & \imgview{6}{12v}\\
16 views        & \imgview{1}{16v}      & \imgview{2}{16v}  & \imgview{3}{16v}   & \imgview{4}{16v}   & \imgview{5}{16v}   & \imgview{6}{16v}\\
20 views        & \imgview{1}{20v}      & \imgview{2}{20v}  & \imgview{3}{20v}   & \imgview{4}{20v}   & \imgview{5}{20v}   & \imgview{6}{20v}\\
GT              & \imgview{1}{gt}       & \imgview{2}{gt}   & \imgview{3}{gt}    & \imgview{4}{gt}    & \imgview{5}{gt}    & \imgview{6}{gt}\\
\bottomrule
\end{tabular}
}
\end{center}
\caption{Examples from the ShapeNet dataset. All predictions are generated by 3D-RETR-B.}
\label{table:shapenet_views}
\end{table}

\section{Model Performance with Different Views}

\begin{figure}
\begin{center}
\includegraphics[width=0.8\linewidth]{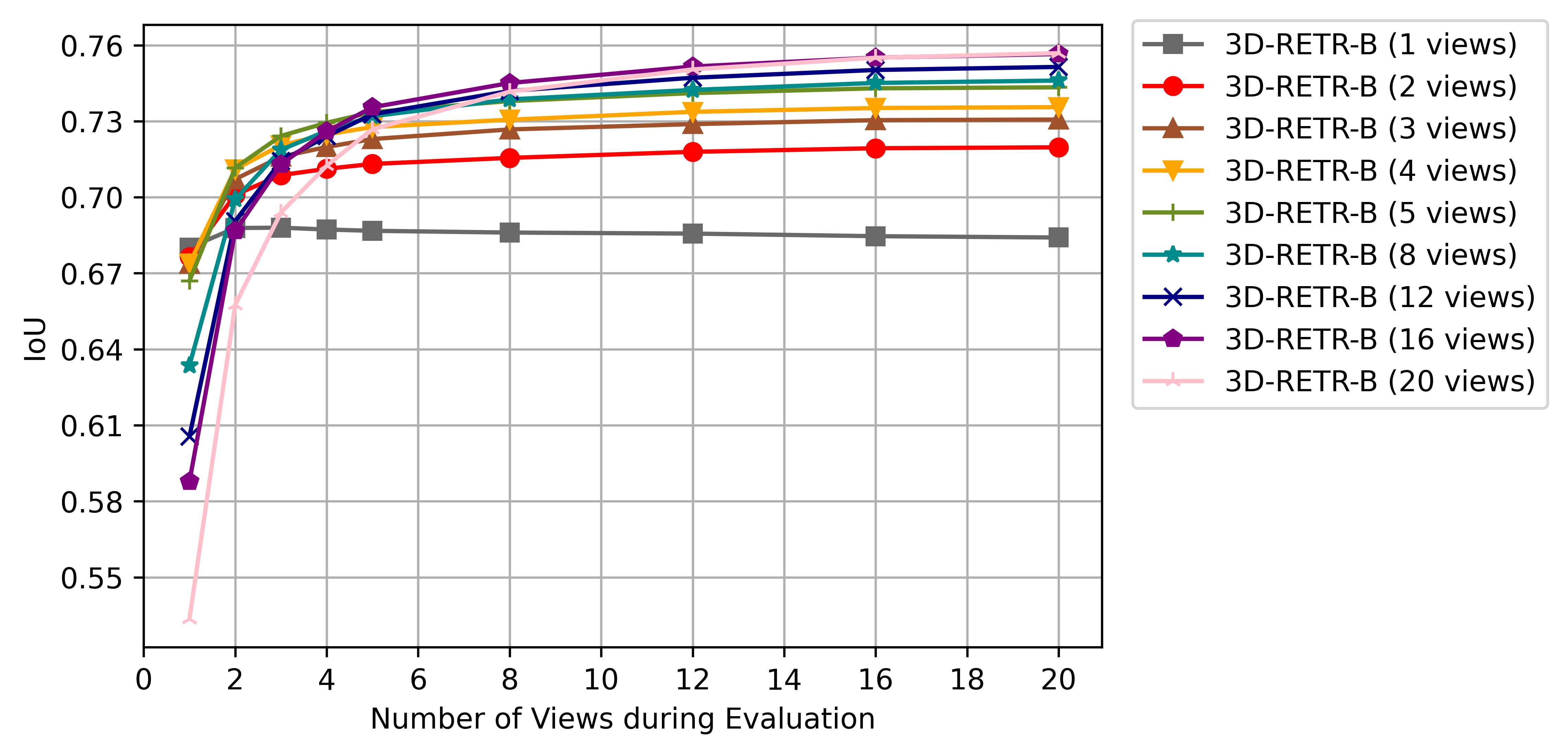}
\end{center}
  \caption{Models performance with different views.}
\label{fig:multi_view}
\end{figure}

In Table 2 of the paper, we show that 3D-RETR trained on three views still outperforms previous state-of-the-art results even when evaluated under different numbers of input views. In Table~\ref{table:different_views} and Figure~\ref{fig:multi_view}, we give additional results on training and evaluating under different numbers of views. We can observe that more views during evaluation can boost model performance.
Another observation is that models trained with more views are not necessarily better than models trained with fewer views, especially when the number of views available during evaluation is far fewer than the number of available views during training. For example, when only one view is available, the model trained with one view reaches an IoU of 0.680, while the model trained with 20 views only reaches an IoU of 0.534.

\newcommand{\p}[1]{\textbf{#1}}

\begin{table}[!ht]
\resizebox{\textwidth}{!}{
\begin{tabular}{lccccccccc}
\toprule
\diagbox{Train}{Eval}                & 1 view   & 2 views & 3 views & 4 views & 5 views & 8 views & 12 views & 16 views & 20 views \\
\midrule\midrule
1 view   & \p{0.680}& 0.688   & 0.688   & 0.687   & 0.687   & 0.686   & 0.686    & 0.685    & 0.684    \\
2 views  & 0.676    & 0.701   & 0.709   & 0.711   & 0.713   & 0.716   & 0.718    & 0.719    & 0.720    \\
3 views  & 0.674    & 0.707   & 0.716   & 0.720   & 0.723   & 0.729   & 0.729    & 0.730    & 0.731    \\
4 views  & 0.674    & 0.711   & 0.721   & 0.725   & 0.728   & 0.731   & 0.734    & 0.735    & 0.736    \\
5 views  & 0.667    &\p{0.712}&\p{0.724}&\p{0.729}& 0.734   & 0.738   & 0.741    & 0.743    & 0.743    \\
8 views  & 0.634    & 0.699   & 0.719   & 0.726   & 0.732   & 0.739   & 0.742    & 0.745    & 0.746    \\
12 views & 0.606    & 0.691   & 0.714   & 0.724   & 0.733   & 0.742   & 0.747    & 0.750    & 0.751    \\
16 views & 0.588    & 0.687   & 0.713   & 0.726   &\p{0.735}&\p{0.745}&\p{0.752} & \textbf{0.755}    & \textbf{0.757}    \\
20 views & 0.534    & 0.657   & 0.694   & 0.712   & 0.727   & 0.742   & 0.750    &\p{0.755} &\p{0.757} \\
\bottomrule
\end{tabular}
}
\vspace{0.2cm}
\caption{Model performance with different views during training and evaluation. \textbf{Bold} indicates the best performance in an evaluation setting.}
\label{table:different_views}
\end{table}

\end{document}